\pdfoutput=1

\documentclass[11pt]{article}

\usepackage[review]{acl}

\usepackage{times}
\usepackage{latexsym}

\usepackage[T1]{fontenc}

\usepackage[utf8]{inputenc}

\usepackage{microtype}

\usepackage{amsmath}
\usepackage{multirow}
\usepackage{graphicx}
\usepackage{amsfonts}
\usepackage{dsfont}
\usepackage{enumitem}
\usepackage{diagbox}
\usepackage{booktabs}
\usepackage[normalem]{ulem}

\DeclareMathOperator*{\argmax}{arg\,max}

\newcommand{\tabitem}{~~\llap{\textbullet}~~}

\title{DimonGen: Diversified Generative Commonsense Reasoning for Explaining Concept Relationships}

\author{Chenzhengyi Liu$^*$ $\quad$ Jie Huang$^*$$^\dag$ $\quad$ Kerui Zhu $\quad$ Kevin Chen-Chuan Chang \\
University of Illinois at Urbana-Champaign, USA \\
 \texttt{\{cl115, jeffhj, keruiz2, kcchang\}@illinois.edu}
}

\begin{document}
\maketitle

\begin{abstract}
In this paper, we propose \textit{DimonGen}, which aims to generate diverse sentences describing concept relationships in various everyday scenarios. To support this, we first create a benchmark dataset for this task by adapting the existing CommonGen dataset. We then propose a two-stage model called MoREE to generate the target sentences. MoREE consists of a mixture of retrievers model that retrieves diverse context sentences related to the given concepts, and a mixture of generators model that generates diverse sentences based on the retrieved contexts. We conduct experiments on the DimonGen task and show that MoREE outperforms strong baselines in terms of both the quality and diversity of the generated sentences. Our results demonstrate that MoREE is able to generate diverse sentences that reflect different relationships between concepts, leading to a comprehensive understanding of concept relationships.\footnote{Code and data are available at \url{https://github.com/liuchenzhengyi/DimonGen}. $^*$Equal contribution. $^\dag$Corresponding author.}
\end{abstract}

\section{Introduction}

Concepts are mental representations of classes or categories of objects, events, or ideas, distinguished by shared characteristics that set them apart from other things. For instance, the concept of ``dog'' represents a class of animals that share characteristics such as being four-legged, having fur, and being domesticated. These concepts are crucial in helping us understand and communicate about the world around us.

To fully grasp concepts, it is important to understand the relationships between them. Researchers have proposed using generated sentences as a means to model these relationships more effectively~\citep{Lin2020CommonGenAC, Huang2022OpenRM, huang2022deer, huang2022ver}. For example, CommonGen~\citep{Lin2020CommonGenAC} aims to generate coherent sentences that describe everyday scenarios involving specific sets of common concepts, while Open Relation Modeling~\citep{Huang2022OpenRM} generates informative sentences that describe relationships between concepts/entities.

However, in real-world scenarios, concepts often refer to broad classes, and their relationships can be complex. This can make it challenging to summarize these relationships through a single sentence. For example, ``dog'' and ``sheep'' are both animal concepts, but while ``dogs'' can herd ``sheep'', they can also attack them. A single sentence would not accurately convey this complexity, leading to an insufficient understanding. Additionally, this approach can also introduce bias, particularly when concepts are related to sensitive topics such as gender or race. For instance, the statement ``\textit{women} are better suited for caregiving roles than \textit{men}.'' is a biased statement.

\begin{figure}[t]
    \centering
    \includegraphics[width=0.45\textwidth]{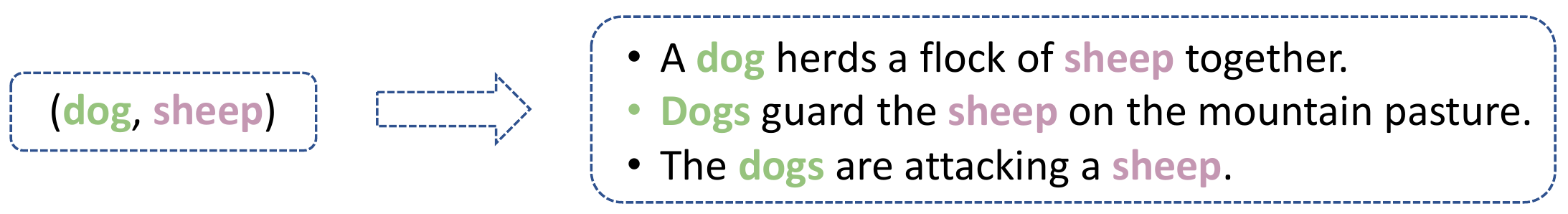}
    \caption{An example of DimonGen. The input is a pair of concepts and the output is a set of sentences that capture different ways in which these concepts interact.}
    \label{fig:example}
\end{figure}

To mitigate the above issues, we propose a new task called \textit{DimonGen: Diversified Generative Commonsense Reasoning}. The task involves generating diverse sentences that describe the relationships between two given concepts, such as the example shown in Fig.~\ref{fig:example} of the concept pair ``dog'' and ``sheep''. This helps build a comprehensive and diverse understanding of the relationships between the concepts in various everyday scenarios.

DimonGen is a challenging task because it requires generating reasonable scenarios for a given pair of concepts without any context. This requires a deep understanding of relational and commonsense knowledge about the concepts. Additionally, the target outputs must reflect diverse relationships between the input concepts. Previous approaches to generating diverse content have used sampling from a designed vocabulary distribution~\citep{Holtzman2020TheCC, Meister2022LocallyTS, Fan2018HierarchicalNS} or encoding inputs to various latent variables~\citep{Zhao2017LearningDD, Cao2020DivGANTD}. However, these methods introduce diversity \textit{only} at the generation stage which may not be suitable for the DimonGen task as it relies on the semantic information from the input contexts.

To overcome the challenges, we propose \textit{\textbf{MoREE}}: \textit{\textbf{M}ixture \textbf{o}f \textbf{R}etrieval-\textbf{E}nhanced \textbf{E}xperts}, a two-stage method that utilizes external knowledge to generate diverse relationship sentences. In the first stage, MoREE retrieves diverse context sentences related to the given concepts using a mixture of retrievers model based on the Mixture of Experts (MoE) model~\cite{Shen2019MixtureMF}. In the second stage, MoREE generates diverse relationship sentences conditioned on the retrieved contexts using a mixture of generators model. An Expectation-Maximization (EM) based matching algorithm is proposed to combine the two stages. By extracting diverse contexts from corporas before generation, MoREE aims to improve the diversity and quality of the generated relationship sentences.

We build a benchmark dataset for DimonGen by adapting the existing CommonGen benchmark~\citep{Lin2020CommonGenAC} and conduct both quantitative and qualitative experiments on the dataset. The results indicate that our proposed MoREE model outperforms well-designed baselines in terms of both the quality and diversity of the generated sentences. For example, in the automatic evaluation, our method gains over $2\%$ in the \textit{BLEU-4}  score for quality and around $5\%$ in \textit{Self-BLEU-4} for diversity. And in our human evaluation, the annotated score (up to $5$) for quality increases from 3.77 to 4.21, and for diverse increases from 3.65 to 3.94.
We also conduct detailed ablation studies and case studies to further verify the effectiveness of our proposed method. Overall, the results suggest that MoREE can generate diverse sentences that reflect relationships between concepts from multiple and varied perspectives.

\section{DimonGen: Diversified Generative Commonsense Reasoning}

We propose a task called \textit{DimonGen} that aims to generate diverse sentences that describe the relationships between a pair of concepts from different perspectives. The task is defined as a diverse constrained text generation task, where the input is a pair of concepts (i.e., $\boldsymbol{x} = \{\boldsymbol{e}_a, \boldsymbol{e}_b\}$) and the output is a set of sentences $\mathcal{Y} = \{\boldsymbol{y}_1, \dots, \boldsymbol{y}_n\}$ that include both concepts and are diverse in terms of their content~(an example is illustrated in Fig. \ref{fig:example}). 

To solve the above task, we propose a two-stage retrieval-enhanced method named MoREE, which consists of a mixture of retriever and generator models~(Fig.~\ref{fig:model}). This method is based on the Mixture of Experts (MoE) model, which will be reviewed in the following section before introducing MoREE.

\subsection{Base Model: Mixtrue of Experts for Diverse Text Generation}\label{sec:moe}

The Mixture of Experts (MoE) is an ensemble technique that was originally designed to increase the capacity of a model~\cite{Jacobs1991AdaptiveMO, Jordan1994HierarchicalMO}. It consists of several expert models that share the same network architecture, but have different probabilities of being assigned to the same training examples. This means that each expert model is exposed to a different subset of the training data, and the MoE ensemble combines them to achieve optimal performance.

In recent years, MoE has been adapted for text-generation tasks to improve diversity in the generation stage~\cite{Shen2019MixtureMF, Cho2019MixtureCS}. Since the mixture base models are trained on different subsets of the training data, they can learn different aspects of the input, leading to a diverse set of generations during the inference phase. Formally, for each training example $(\boldsymbol{x}, \boldsymbol{y})$ where $\boldsymbol{y} \in \mathcal{Y}$ is a relation description, if there are $n$ expert models with a set of latent variables $\mathcal{Z} = \{ \boldsymbol{z}_1, \dots, \boldsymbol{z}_n \}$ as identifiers, the likelihood of the MoE model is formulated as the following marginal likelihood:
\begin{equation}
    p(\boldsymbol{y} | \boldsymbol{x}; \theta) = \sum_{i=1}^{n} p(\boldsymbol{z}_i| \boldsymbol{x}; \theta) p(\boldsymbol{y} | \boldsymbol{z}_i, \boldsymbol{x}; \theta),
\end{equation}
where $\theta$ represents the model weights. 

To promote diversity among the different expert models, the training examples are split into subsets with distinct elements, and each expert model is trained on one subset. This training process is done through a hard-EM algorithm as follows~\cite{Shen2019MixtureMF, Yu2022DiversifyingCG}:

\begin{itemize}[leftmargin=*]
    \item \textbf{E-step:} for each training example $(\boldsymbol{x}, \boldsymbol{y})$, select the expert model $z_i\in\mathcal{Z}$ that maximizes the posterior probability $p(\boldsymbol{z}_i|\boldsymbol{x}, \boldsymbol{y}; \theta)$ using current model weights $\theta$ with the equation $\boldsymbol{z}_i = \argmax_{\boldsymbol{z}\in\mathcal{Z}} p(\boldsymbol{y}, \boldsymbol{z}|\boldsymbol{x}; \theta)$.
    \item \textbf{M-step:} update the model weights $\theta$ through the gradients $\nabla_\theta \log p(\boldsymbol{y}, \boldsymbol{z}_i | \boldsymbol{x}; \theta)$ of selected expert model $\boldsymbol{z}_i$.
\end{itemize}

The hard-EM algorithm is performed by iterating these two steps. It should be noted that this algorithm can be easily applied to a batch learning algorithm by updating the model weights for each batch during the M-step. Finally, by assuming a uniform prior of expert models, the loss function could be formulated as
\begin{equation}
    \mathcal{L} = \mathbb{E}_{(\boldsymbol{x}, \boldsymbol{y})} [\min_i -\log p(\boldsymbol{y}|\boldsymbol{z}_i, \boldsymbol{x}; \theta)].
\end{equation}

\subsection{MoREE: Mixture of Retrieval-Enhanced Experts}

The DimonGen task poses a significant challenge as it requires relational commonsense reasoning and the generation of diverse content with the minimal input information. Traditional methods for encouraging diverse text generation focus on introducing diversity in the generation stage through diversified decoding or sampling mechanisms~\citep{Meister2022LocallyTS, Fan2018HierarchicalNS, Zhao2017LearningDD, Cao2020DivGANTD}. However, these methods are not suitable for the DimonGen task due to the limited input information and the need for diversified relational reasoning. Our experiments in Sec.~\ref{sec:res} show that even with powerful pre-trained language models, these methods struggle to solve this task.

\begin{figure*}[t]
    \centering
    \includegraphics[width = \textwidth]{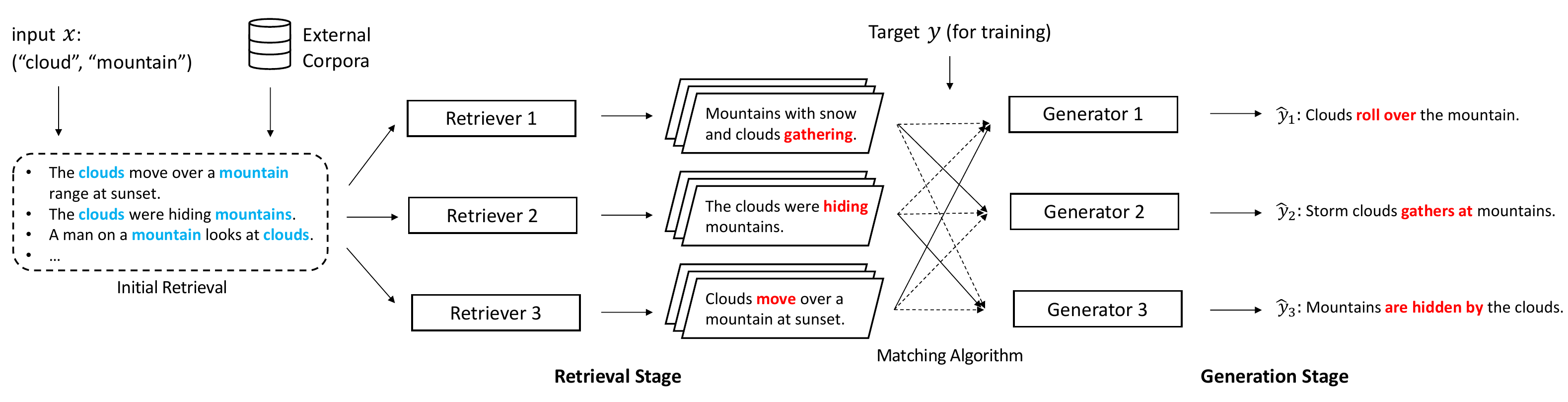}
    \caption{The overall framework of the proposed MoREE method, which includes two stages: 1) retrieval stage with a mixture of retrievers model to extract diversified contexts, 2) generation stage with a mixture of generators model. And a matching algorithm is used to concatenate these two stages.}
    \label{fig:model}
\end{figure*}

To address this challenge, we propose a diversified retrieval-enhanced method named \textbf{M}ixture \textbf{o}f \textbf{R}etrieval-\textbf{E}nhanced \textbf{E}xperts (MoREE). Our overall framework is illustrated in Fig.~\ref{fig:model} which consists of two stages. In the first stage, we use a mixture of retrievers model to extract several sets of diverse context sentences as auxiliary inputs to help with the generation process. In the second stage, we use a mixture of generators model to generate diverse outputs and propose a matching algorithm to assign the appropriate contexts to the target outputs.

\subsubsection{Retrieval Stage}\label{sec:ret}

To better understand the relationships between given concepts, we introduce a retrieval stage to gather context sentences from external corpora $\mathcal{C}$. Given an input concept pair $\boldsymbol{x} = \{\boldsymbol{e}_a, \boldsymbol{e}_b\}$, we aim to retrieve several diversified sets of relational contexts $\{\mathcal{S}_1, \dots, \mathcal{S}_n\}$, where $\mathcal{S}_i = \{\boldsymbol{s}_1^i, \dots, \boldsymbol{s}_k^i\}$ is a set of context sentences containing $\boldsymbol{x}$.

We train the retriever models on a binary classification task. Given a candidate sentence from external knowledge corpora $\boldsymbol{s}_j \in \mathcal{C}$, we concatenate it with the input $\boldsymbol{x}$ and use it as input:

\begin{equation}
\begin{gathered}
\boldsymbol{x}_j^{\text{re}} = \text{ [CLS] } \boldsymbol{x}  \text{ [SEP] } \boldsymbol{s}_j \text{ [SEP] }\label{con_input} \\
    \boldsymbol{x} = \boldsymbol{e}_a \text{ [SEP] } \boldsymbol{e}_b,
\end{gathered}
\end{equation}
where [CLS] and [SEP] are special tokens in pre-trained language models. The model's task is to predict a label $y_c$ from $[0, 1]$, indicating the confidence of the candidate sentence being a true relational context for the input concepts. 
For each input, we use its target output sentences in the dataset as positive examples and randomly sample the same number of negative examples from its retrieved candidate sentences.

To extract diversified contexts for each input concept pair, we introduce the mixture of experts (MoE) method into the retriever model. Since independently parameterizing each expert may cause an overfitting problem, we follow the weight-sharing schema in~\citet{Shen2019MixtureMF} with a unique identifier to solve this issue. To make the MoE models more easily understood by pre-trained language models, for each expert model, we design its unique identifier as latent variables $\boldsymbol{z}_i = z^i_1, \dots, z^i_m$ which is a randomly sampled prefix token sequence in the model vocabulary. Once an expert is chosen, we could train the model by concatenating the latent variable and input concepts with contexts as the final input:
\begin{equation}
    \boldsymbol{x}_{ji}^{\text{re}} = \boldsymbol{z}_i \text{ [CLS] } \boldsymbol{x}  \text{ [SEP] } \boldsymbol{s}_j \text{ [SEP] }.  \label{fianl_input}
\end{equation}

We apply the hard-EM algorithm~\cite{Shen2019MixtureMF, Yu2022DiversifyingCG} to train our mixture of retrievers model. For each iteration, at E-step, we assign the expert model to each input; at M-step, we update all the expert models with the assigned inputs. With this process, the total training loss turns into an expectation form: 
\begin{equation}
    \mathcal{L}_c = \mathbb{E}_{(\boldsymbol{x}_j^{\text{re}}, y_c)} [\min_i -\log p(y_c| \boldsymbol{x}_{ji}^{\text{re}}; \theta)].
\end{equation}

However, during experiments, we find the binary classification problem has obvious patterns, and simply applying the hard-EM algorithm may lead to a severe overfitting problem (i.e., one expert always predicts one class label). To solve this problem, we propose a regularization term based on Jenson Shannon divergence~\cite{Sibson1969InformationR} to penalize the output probability distribution over different labels among experts. Given the output probability distribution of $n$ experts $\{P_1, \dots, P_n\}$, the regularization loss is calculated as an average of the Kullback-Leibler (KL) distances between each distribution and the distribution center:
\begin{equation}
    \mathcal{L}_r = \frac{1}{n} \sum^{n}_{i=1} D_{\text{KL}} (P_i || \frac{1}{n} \sum_j P_j),
\end{equation}
where $D_{\text{KL}}(\cdot || \cdot)$ is the KL divergence. 

The final loss function for our mixture of retrievers model is a weighted sum of the two:
\begin{equation}
    \mathcal{L} = \mathcal{L}_c + \alpha \mathcal{L}_r,
\end{equation}
where $\alpha$ is a hyperparameter  to balance the two losses.

\subsubsection{Generation Stage}

At the generation stage, we fine-tune a mixture of generators model to generate qualified and diversified relationship sentences with retrieved sets of contexts. Given an input concept pair $\boldsymbol{x} = \{\boldsymbol{e}_a, \boldsymbol{e}_b\}$ and several diversified sets of context sentences $\{\mathcal{S}_1, \dots, \mathcal{S}_n\}$ from the retrieval stage, our goal is to generate a set of relationship sentences $\hat{\mathcal{Y}} = \{ \hat{\boldsymbol{y}}_1, \dots, \hat{\boldsymbol{y}}_n \}$. 

For each input concept pair $\boldsymbol{x}$ and each set of its context sentences  $\mathcal{S}_i = \{\boldsymbol{s}_1^i, \dots, \boldsymbol{s}_k^i\}$, we concatenate all of their input token sequences with the expert's latent variable $\boldsymbol{z}_i$ to construct the final input as
\begin{equation}
    \boldsymbol{x}_{i}^{\text{gen}} = \boldsymbol{z}_i \text{ [CLS] } \boldsymbol{x} \text{ [SEP] } \boldsymbol{s}_1^i \text{ [SEP] } \dots \boldsymbol{s}_k^i.
\end{equation}
By applying the same method for all sets of retrieved sentences, we can obtain $n$ different context-aware inputs $\{\boldsymbol{x}_{1}^{\text{gen}}, \dots, \boldsymbol{x}_{n}^{\text{gen}}\}$.

However, the retrieved contexts are not present in the original dataset and thus, there is no explicit link between the target outputs and the retrieved contexts. To address this issue, we propose a matching algorithm based on a hard-EM algorithm similar to the one used in the MoE process. For each input, we evaluate its compatibility with each target output by calculating the posterior probability $p(\boldsymbol{y}_j| \boldsymbol{x}_{i}^{\text{gen}}; \theta)$ using the current generator model's parameters $\theta$. The context-aware input is then assigned to the target output with the highest score:
\begin{equation}
    \boldsymbol{y}_{i}^{\text{target}} = \argmax_{\boldsymbol{y}_j \in \mathcal{Y}} \quad p(\boldsymbol{y}_j| \boldsymbol{x}_{i}^{\text{gen}}; \theta).\label{eq:assign}
\end{equation}

In the training phase, we use Eq.~\eqref{eq:assign} to construct training examples at E-step and then use these examples to fine-tune a mixture of generators model at M-step. In the inference phase, for each input, we feed all the diversified context-aware inputs into the generator model to generate diverse results.

\section{Experiments}

\subsection{Dataset Construction and Analysis}

We construct our DimenGen benchmark dataset by combining the CommonGen dataset~\cite{Lin2020CommonGenAC}, which contains high-quality descriptive sentences for everyday relations between input concepts, and ConceptNet~\cite{Speer2017ConceptNet5A}, a semantic graph with nodes representing concepts and edges indicating the category of the relationship between them. To build our dataset, we first cluster all pairs of input concepts present in the CommonGen dataset and collect their corresponding relational sentences as target relationship sentences. We then verify the informativeness and correctness of the dataset using ConceptNet. Specifically, we ensure that each concept set in every target relationship sentence contains a \textit{path} between the input concept pair on ConceptNet, verifying the existence of a semantic relationship between the concepts described by a chain of category names. This approach helps us establish the semantic relationships between the input concepts in a systematic manner and ensures that the generated sentences contain coherent and meaningful relationships.

To encourage diversity, given a target set of generations for an input concept pair, we first embed each target sentence into latent space with Sentence-BERT~\cite{Reimers2019SentenceBERTSE} model and calculate the cosine similarity for each pair of them. Next, we filter out the generations that have pair-wise cosine similarity higher than a pre-set threshold $p=0.75$ in our experiments. For each input concept pair, we limit its target references within $3 \sim 5$. This is because if the number is too small, it is difficult for the model to learn the diversity in the references, while if the number is too large, the models will be trained in a biased manner towards some input concept pairs.

To help evaluate the generalization ability, following CommonGen~\cite{Lin2020CommonGenAC}, we explicitly control the ratio of unseen concept compositions between input concepts in test examples and target outputs in training examples. 
Table~\ref{tab:static} shows the basic statistics of the dataset. We totally extract $16212$ examples in our dataset with $15263$, $665$, and $1181$ split for training, dev, and test. The ratio of unseen concept compositions is $92\%$ and $98\%$ for dev and test respectively. The highly unseen concept compositions make the DimonGen task a difficult problem to solve, which requires the model to be capable of generalized reasoning ability.

\begin{table}[tp]
\centering
\scalebox{0.9}{
\begin{tabular}{c|c|c|c}
\hline
                  & train  & dev   & test  \\ \hline
Number   & 15,263 & 665   & 1,181 \\ \hline
Unseen ratio (\%) & -      & 91.73 & 98.31 \\ \hline
Avg. ref. number & 4.13     & 3.71 & 3.38 \\ \hline
\end{tabular}
}

\centering
\scalebox{0.9}{
\begin{tabular}{c|c|c|c}
\hline
                    & $3$-targets & $4$-targets & $5$-targets \\ \hline
ratio (\%) & 34.76     & 24.16     & 41.07         \\ \hline
\end{tabular}
}
\caption{The statistics of the DimonGen dataset.} \label{tab:static}
\end{table}

For the diversity, the average numbers of target relationship sentences for each example are $4.13$, $3.71$, and $3.38$ for training, dev, and test sets respectively. It is also noted that there are over $41\%$ examples that have $5$ target outputs. The high ratio of examples with $5$ target references not only contributes to increasing the models' ability to generate diverse outputs but also helps to build comprehensive evaluation metrics.

\subsection{Experimental Setup}

\paragraph{Baselines.} Since we are targeting the DimonGen task with many references for each input, we compare with several strong baseline models with diverse text generation capabilities. Generally, previous works introduce the diversity at the generation stage by either sampling the next word by a probability distribution (Sampling-based methods) or incorporating mixture components in the generator model (MoE-based methods). Different from previous works, our MoREE model introduces diversity by extracting the diverse contexts from the external corpora at the retrieval stage. 

\begin{itemize}[leftmargin=*]
    \item \textbf{Sampling-based methods.} Sampling methods create diverse outputs at the inference phase of the generation stage. These methods sample the next token with a designed probability distribution of the vocabulary, rather than simply maximizing the likelihood. We compare with three strong sampling-based methods: Top-k sampling~\cite{Fan2018HierarchicalNS} truncates the sampling pool by keeping only the top-k candidates for each token in the generation. Top-p sampling~\cite{Holtzman2020TheCC} cuts off the next-token sampling pool from a threshold of the probability mass. Typical sampling~\cite{Meister2022LocallyTS} constrains the generated words to expected information content by shifting the truncation set with a conditional entropy of prior content.
    
    \item \textbf{MoE-based methods.} MoE-based methods introduce diversity at the training phase of the generation stage by using diverse latent variables. We compare with two of them: MoE~\cite{Shen2019MixtureMF} is the vanilla MoE model for diverse text generation we discussed in Sec.~\ref{sec:moe}. MoKGE~\cite{Yu2022DiversifyingCG} incorporates commonsense knowledge from an external graph and uses the MoE model to generate diverse outputs. Compared to our model, MoKGE also extracts information from external knowledge, but it only introduces diversity at the generation stage.
\end{itemize}

\begin{table*}[tp]
\centering
\scalebox{0.76}{
\begin{tabular}{|cl|ccc|cc|cc|}
\hline
\multicolumn{2}{|c|}{\multirow{2}{*}{\textbf{Method}}}                                                                                 & \multicolumn{3}{c|}{\textbf{Quality (top-k) $\uparrow$}}                                               & \multicolumn{2}{c|}{\textbf{Pairwise diversity $\downarrow$}}     & \multicolumn{2}{c|}{\textbf{Corpus diversity $\uparrow$}}      \\ \cline{3-9} 
\multicolumn{2}{|c|}{}                                                                                                                 & \multicolumn{1}{p{1.8cm}|}{\centering BLEU-4}         & \multicolumn{1}{p{1.8cm}|}{\centering ROUGE-l}        & \multicolumn{1}{p{1.8cm}|}{\centering S. R.} & \multicolumn{1}{p{1.8cm}|}{\centering self-B.-4}    & \multicolumn{1}{p{1.8cm}|}{\centering self-R.-l}   & \multicolumn{1}{p{1.8cm}|}{\centering Entropy-4}    & \multicolumn{1}{p{1.8cm}|}{\centering Distinct-4}    \\ \hline
\multicolumn{1}{|c|}{\multirow{3}{*}{\textbf{\begin{tabular}[c]{@{}c@{}}Sampling\\ methods\end{tabular}}}} & \textbf{Top\_k sampling}  & \multicolumn{1}{c|}{14.97}          & \multicolumn{1}{c|}{40.29}          & 87.75           & \multicolumn{1}{c|}{38.54}          & 61.27          & \multicolumn{1}{c|}{9.50}          & 74.53          \\ \cline{2-9} 
\multicolumn{1}{|c|}{}                                                                                     & \textbf{Top\_p sampling}  & \multicolumn{1}{c|}{15.35}          & \multicolumn{1}{c|}{40.17}          & 87.30           & \multicolumn{1}{c|}{33.58}          & 56.57          & \multicolumn{1}{c|}{9.60}          & 78.22          \\ \cline{2-9} 
\multicolumn{1}{|c|}{}                                                                                     & \textbf{Typical sampling} & \multicolumn{1}{c|}{15.26}          & \multicolumn{1}{c|}{40.42}          & 87.60           & \multicolumn{1}{c|}{35.05}          & 57.99          & \multicolumn{1}{c|}{9.58}          & 77.36          \\ \hline
\multicolumn{1}{|c|}{\multirow{3}{*}{\textbf{\begin{tabular}[c]{@{}c@{}}MoE\\ methods\end{tabular}}}}      & \textbf{MoE}              & \multicolumn{1}{c|}{16.70}          & \multicolumn{1}{c|}{40.88}          & 87.84           & \multicolumn{1}{c|}{30.86}          & 51.16          & \multicolumn{1}{c|}{9.49}          & 75.87          \\ \cline{2-9} 
\multicolumn{1}{|c|}{}                                                                                     & \textbf{MoKGE}            & \multicolumn{1}{c|}{16.60}          & \multicolumn{1}{c|}{41.34}          & 88.37           & \multicolumn{1}{c|}{29.73}          & 50.02          & \multicolumn{1}{c|}{9.58}          & 79.12          \\ \cline{2-9} 
\multicolumn{1}{|c|}{}                                                                                     & \textbf{MoREE (ours)}     & \multicolumn{1}{c|}{\textbf{19.06}} & \multicolumn{1}{c|}{\textbf{43.17}} & \textbf{91.69}  & \multicolumn{1}{c|}{\textbf{24.85}} & \textbf{46.85} & \multicolumn{1}{c|}{\textbf{9.70}} & \textbf{83.62} \\ \hline
\end{tabular}
}
\caption{Results of DimonGen task for different methods; evaluation metrics contain three dimensions; ``S.R.'', ``self-B.-4'', and ``self-R.-l'' are abbreviations for ``Successful Rate'', ``self-BLEU-4'', and ``self-ROUGE-l'' respectively (note the lower the pairwise diversity score ``$\downarrow$'', the better the performance on diversity).} \label{tab:result}
\end{table*}

\paragraph{Implementation.} 
In our proposed method, we utilize external corpora from VATEX~\cite{Wang2019VaTeXAL}, ActivityNet~\cite{Krishna2017DenseCaptioningEI}, SNLI~\cite{Bowman2015ALA}, and MNLI~\cite{Williams2017ABC} for retrieval purposes. These datasets comprise high-quality descriptive sentences and are widely employed in commonsense benchmarking tasks. We retrieve all sentences containing both input concepts to create a candidate pool. In cases where there are insufficient candidates, we substitute the concepts with the smallest cosine similarity in each sentence, according to their Word2Vec~\cite{Mikolov2013EfficientEO} embeddings.
We use pre-trained Roberta models~\cite{Liu2019RoBERTaAR} as its base model to rank and select the candidates in the retrieval stage. 
For the generation stage, we use the pre-trained BART model~\cite{Lewis2019BARTDS} as the base model for all baseline methods and our proposed method for a fair comparison.  We require each method to generate $k=3$ relationship sentences in our experiments because the minimum reference sentences’ number in the dataset is $3$.

We use Huggingface's Transformers~\cite{Wolf2019TransformersSN} to implement the code and perform a grid search to find the best hyper-parameters for all baseline methods. Our models were trained by one NVIDA RTX A40 GPU card with about 4-5 hours of training on the DimonGen dataset.

\paragraph{Metrics.} To evaluate the performance of our proposed DimonGen task, we use three different evaluation metrics: quality, pairwise diversity, and corpus diversity.

\begin{itemize}[leftmargin=*]
    \item \textbf{Quality metrics.} For quality evaluation, we use both $N$-gram-based metrics such as \textit{BLEU}~\cite{Papineni2002BleuAM} and \textit{ROUGE}~\cite{Lin2004ROUGEAP}, as well as the concept overlapping rate (\textit{Success Rate}) between the input and generated sentences. We make a slight modification for the DimonGen task by first requiring the model to generate a set of top-$k$ candidates, then evaluating the quality between each generated candidate and the target references. The best candidate with the highest score is chosen and its score is used for the quality metrics.
    
    \item \textbf{Pairwise diversity.} To measure pairwise diversity, we compute the average score of $N$-gram-based evaluation metrics between all pairs of generations in the generated candidate set. The lower the average score is, the higher the evaluated pairwise diversity will be. These metrics are named \textit{Self-BLEU} and \textit{Self-ROUGE}~\cite{Zhu2018TexygenAB}.
    
    \item \textbf{Corpus diversity.} To evaluate the corpus diversity of the generated text, we use two widely-used metrics: \textit{Distinct-$n$}\cite{Li2015ADO} and \textit{Entropy-$n$}\cite{Zhang2018GeneratingIA}. \textit{Distinct-$n$} is computed by taking the ratio of the number of unique $n$-grams to the total number of $n$-grams in the generated sentences. On the other hand, \textit{Entropy-$n$} calculates the average uncertainty of the $n$-gram distribution within one generation, providing an estimate of the diversity of the generated text.
    
\end{itemize}

\subsection{Experimental results}\label{sec:res}

\begin{table*}[tp]
\centering
\scalebox{0.76}{
\begin{tabular}{|l|ccc|cc|cc|}
\hline
\multicolumn{1}{|c|}{\multirow{2}{*}{\textbf{Method}}} & \multicolumn{3}{c|}{\textbf{Quality (top-k) $\uparrow$}}                                              & \multicolumn{2}{c|}{\textbf{Pairwise diversity $\downarrow$}}     & \multicolumn{2}{c|}{\textbf{Corpus diversity $\uparrow$}}      \\ \cline{2-8} 
\multicolumn{1}{|c|}{}                                   & \multicolumn{1}{p{1.8cm}|}{\centering BLEU-4}         & \multicolumn{1}{p{1.8cm}|}{\centering ROUGE-l}        & \multicolumn{1}{p{1.8cm}|}{\centering S. R.} & \multicolumn{1}{p{1.8cm}|}{\centering self-B.-4}    & \multicolumn{1}{p{1.8cm}|}{\centering self-R.-l}   & \multicolumn{1}{p{1.8cm}|}{\centering Entropy-4}    & \multicolumn{1}{p{1.8cm}|}{\centering Distinct-4}    \\ \hline
\textbf{MoREE}                                  & \multicolumn{1}{c|}{\textbf{19.06}} & \multicolumn{1}{c|}{\textbf{43.17}} & 91.69          & \multicolumn{1}{c|}{\textbf{24.85}} & \textbf{46.85} & \multicolumn{1}{c|}{\textbf{9.70}} & \textbf{83.62} \\ \hline
\quad \textbf{w/o mixture of retrievers}                                  & \multicolumn{1}{c|}{16.91}          & \multicolumn{1}{c|}{41.84}          & 91.04          & \multicolumn{1}{c|}{27.77}          & 50.43          & \multicolumn{1}{c|}{9.54}          & 80.69          \\ \hline
\quad \textbf{w/o regularization term}                                & \multicolumn{1}{c|}{18.57}          & \multicolumn{1}{c|}{42.88}          & \textbf{91.87} & \multicolumn{1}{c|}{29.40}          & 51.45          & \multicolumn{1}{c|}{9.55}          & 79.31          \\ \hline
\quad \textbf{w/o matching algorithm}                                 & \multicolumn{1}{c|}{16.64}          & \multicolumn{1}{c|}{41.78}          & 91.27          & \multicolumn{1}{c|}{28.98}          & 50.96          & \multicolumn{1}{c|}{9.48}          & 77.47          \\ \hline
\end{tabular}
}
\caption{Ablation study of our proposed method by taking off each component: 1) mixture of retrievers, 2) regularization term, 3) matching algorithm respectively.} \label{tab:abla}
\end{table*}

\begin{table}[tp]
\centering
\scalebox{0.8}{
\begin{tabular}{lccc}
\toprule
\textbf{Method} & \textbf{Quality} & \textbf{Diversity} & \textbf{Gra \& Flu} \\
\midrule
\textbf{DimonGen} & \textbf{4.70} & \textbf{4.25} & \textbf{4.67} \\
\midrule
\textbf{Typical} & 3.65 & 3.12 & 4.35 \\
\textbf{MoKGE} & 3.77 & 3.65 & \textbf{4.63} \\
\textbf{MoREE (ours)} & \textbf{4.21} & \textbf{3.94} & 4.61 \\
\bottomrule
\end{tabular}
}
\caption{Qualitative evaluation results of the DimonGen dataset and generations from three different methods} \label{tab:hum}
\end{table}

The experimental results in Table~\ref{tab:result} show that our proposed MoREE model outperforms all five baseline models in both quality and diversity metrics on the DimonGen task. Specifically, our method achieves a $2\%$ improvement in \textit{BLEU-4} compared to other baseline models in terms of quality, and outperforms the strong baseline MoKGE model by around $5\%$ in \textit{Self-BLEU-4} for pairwise diversity and $4\%$ in \textit{distinct-4} for corpus diversity. These results demonstrate the superior diverse generation capabilities of our proposed method.

Additionally, the results show that MoE-based methods have a significant advantage over sampling-based methods in terms of diversity, with an approximate $5\%$ improvement in \textit{Self-BLEU-4} and $6\%$ in \textit{distinct-4}. Furthermore, retrieving from external corpora improves performance on concept-related evaluation metrics, as shown by the superior \textit{success rate} of the MoKGE model and MoREE model compared to the vanilla MoE model. Our MoREE method specifically achieves a $4\%$ gain in this metric, indicating the effectiveness of the mixture retriever in extracting high-quality contexts to assist diverse generations.

\subsection{Ablation Study}

In order to gain a deeper understanding of our proposed two-stage framework, we conduct an ablation study by removing different components of our method and comparing the results. Specifically, we remove the MoE module from the retrieval stage, remove the proposed regularization term for training MoE retrievers, and replace the EM-based matching algorithm for the generation stage with random selection. Table~\ref{tab:abla} displays the results, revealing the following insights:

\begin{itemize}[leftmargin=*]

    \item For the retrieval stage, employing a \textbf{mixture of retrievers} improves both quality and diversity. When using a single retriever model with MoE generators, the \textit{BLEU-4} score drops from $19.06$ to $16.91$, and the \textit{Self-BLEU-4} score increases from $24.85$ to $27.77$. This suggests incorporating diversity into the retrieval stage with a mixture of retrievers can enhance diverse commonsense reasoning capabilities.
    
    \item For the retrieval stage, the proposed \textbf{regularization term} significantly boosts diversity. Without the regularization term in the loss function during the training process, the \textit{Self-BLEU-4} score increases from $24.85$ to $29.40$. This demonstrates that our proposed regularization term helps the retriever balance the distribution of different models, which in turn improves the diversity of the retrieved contexts and generations.
    
    \item For the generation stage, the proposed \textbf{matching algorithm} greatly enhances both quality and diversity. Specifically, our proposed EM-based matching algorithm for matching retrieved contexts to target output gains over $2\%$ on the \textit{BLEU-4} score and $6\%$ on the \textit{distinct-4} score compared to random selection. This indicates that the matching algorithm can effectively assign appropriate contexts to generations, improving the quality and diversity of the generations.
 
\end{itemize}

\subsection{Human Evaluation}

In order to understand the effectiveness of our proposed MoREE method, we conduct the human evaluation by asking three annotators to assign grades (up to $5$) of the generated relationship sentences. We randomly sample $100$ examples from the test set of the DimonGen dataset and compare our method with the typical sampling and MoKGE methods. Following~\citet{Yu2022DiversifyingCG}, we design three evaluation dimensions: quality, diversity, and grammar \& fluency (gra \& flu). 

The human evaluation results in Tab.~\ref{tab:hum} shows that the DimonGen dataset receives high scores for quality and diversity, indicating that the majority of examples in the dataset are well-written and diverse. Our proposed MoREE method outperforms the two baseline methods in terms of quality and diversity and achieves similar scores for grammar and fluency. This demonstrates that our method is able to effectively capture the complex relationships between concepts in real-world scenarios while also generating a variety of unique and accurate relationship sentences.

\subsection{Case study}

\begin{table*}[tp]
\centering
\scalebox{0.65}{
\begin{tabular}{lll}
\hline
\diagbox{\textbf{Method}}{\textbf{Input}}           & \multicolumn{1}{c}{\textbf{(``dog'',   ``sheep'')}}                                                                                                                                                                       & \multicolumn{1}{c}{\textbf{(``airport'',   ``way'')}}                                                                                                                                                                                                                                 \\ \hline \hline
\textbf{Typical sampling} & \begin{tabular}[c]{@{}l@{}}\tabitem A man is walking along a road with a dog and two sheep.\\ \tabitem A group of sheep and a dog are grazing on the grass.\\ \tabitem A man and a dog are standing in a field with sheep.\end{tabular} & \begin{tabular}[c]{@{}l@{}}\tabitem A plane is on its way to an airport.\\ \tabitem An airplane is making its way down the runway at an airport.\\ \tabitem A motorcade makes its way down the runway at an airport.\end{tabular}                                                                   \\ \hline
\textbf{MoKGE}            & \begin{tabular}[c]{@{}l@{}}\tabitem a dog is eating a sheep.\\ \tabitem Sheep and dogs are grazing in a meadow.\\ \tabitem A dog is walking around a field with sheep.\end{tabular}                                                     & \begin{tabular}[c]{@{}l@{}}\tabitem a woman makes her way through the airport.\\ \tabitem passengers make their way through the airport.\\ \tabitem A woman is making her way through an airport.\end{tabular}                                                                                      \\ \hline
\textbf{MoREE (ours)}            & \begin{tabular}[c]{@{}l@{}}\tabitem The dog is herding sheep with a farmer nearby.\\ \tabitem A dog is chasing a flock of sheep.\\ \tabitem The dog follows the sheep through the gate.\end{tabular}                                    & \begin{tabular}[c]{@{}l@{}}\tabitem passengers at an airport are carrying their luggage to and from \\ the terminal as they make their way to their destinations.\\ \tabitem A plane is on its way to the airport.\\ \tabitem A plane is making its way down the runway at an airport.\end{tabular} \\ \hline
\textbf{DimonGen (Gold)}  & \begin{tabular}[c]{@{}l@{}}\tabitem A dog herds a flock of sheep together.\\ \tabitem dogs guard the sheep on the mountain pasture.\\ \tabitem The dogs are attacking a sheep.\end{tabular}                                             & \begin{tabular}[c]{@{}l@{}}\tabitem People make their way off a plane toward the airport.\\ \tabitem There is a gray and red plane on the run way at the airport.\\ \tabitem US Airways plane moves on a taxi way near its gate at an airport.\end{tabular}                                         \\ \hline
\end{tabular}
}
\caption{Generated examples for input concept pairs (``dog'', ``sheep'') and (``airport'', ``way'') } \label{tab:case}
\end{table*}

Table~\ref{tab:case} illustrates some generation examples for input concept pairs $\{$``dog'', ``sheep''$\}$ and $\{$``airport'', ``way''$\}$ with different methods, which shows that: 

\begin{itemize}[leftmargin=*]
\item For the input pair ``dog'' and ``sheep'', the generations produced by the baseline methods contain some unreasonable outputs, such as ``Sheep and dogs are grazing in a meadow.'' In contrast, our proposed MoREE method generates more reasonable and diverse outputs, such as ``The dog is heading sheep with a farmer nearby'' and ``A dog is chasing a flock of sheep.''

\item For the input pair ``airport'' and ``way'', the baseline methods tend to generate plain and repetitive outputs, such as ``Passengers make their way through the airport.'' In contrast, our proposed MoREE method can accurately capture the relationships between concepts, for example, ``A plane is on its way to the airport.''

\end{itemize}

\section{Related Work}

\paragraph{Generative relational reasoning} attempts to generate a coherent sentence involving a pair or a set of concepts/entities~\citep{Lin2020CommonGenAC, Huang2022OpenRM, huang2022deer, huang2022ver}.
For instance, \citet{Lin2020CommonGenAC} introduce \textit{CommonGen}, which aims to generate a coherent sentence that describes an everyday scenario involving a given set of common concepts. \citet{Huang2022OpenRM} propose \textit{Open Relation Modeling}, which aims to generate an informative sentence describing relationships between concepts. 
However, these methods do not consider the diversity of possible relationships that can exist between concepts, leading to a limited understanding of relationships between the concepts.

\paragraph{Incorporating diversity at inference phrase} is achieved by sampling methods. Instead of selecting the next token based on maximum likelihood~\citep{Freitag2017BeamSS}, tokens are sampled from a probability distribution of the vocabulary. For example, \citet{Fan2018HierarchicalNS} reduces the sampling pool by keeping only the top-k candidates for each token in the generation. \citet{Holtzman2020TheCC} limits the next-token sampling pool by a threshold of the probability mass. \citet{Meister2022LocallyTS} restrict the generated words to expected information content by shifting the truncation set with a conditional entropy of prior content. While these methods reduce the training effort for neural models, they are criticized for the low quality of generations~\citep{Zhang2020TradingOD}.

\paragraph{Incorporating diversity at the training phase} is achieved through diverse model structures. Specifically, \citet{Zhao2017LearningDD} propose a conditional variational autoencoder-based framework to embed each input into a latent distribution. \citet{Cao2020DivGANTD} construct their model based on a conditional generative adversarial network with a diversity loss term. \citet{Shen2019MixtureMF} and \citet{Cho2019MixtureCS} utilize a mixture of experts model to encourage diverse outputs from different expert models. Among previous works, \citet{Yu2022DiversifyingCG}'s method is most similar to ours. They propose to first extract commonsense knowledge from external knowledge graphs and then use an MoE model to generate diverse outputs. While this work considers incorporating external knowledge to improve generation quality, it falls short in increasing diversity due to the naive retriever shared among all the generators.

\section{Conclusion}

While previous approaches have used generated sentences to model concept/concept relationships, these methods often rely on a single sentence and can be insufficient or biased in conveying the complexity of these relationships. To address this issue, we propose DimonGen, a task for generating diverse sentences that describe concept relationships in various everyday scenarios. 
To solve the proposed task,
we design a two-stage model called MoREE, which combines a mixture of retriever and generator models. Our experimental results demonstrate the effectiveness of MoREE in generating coherent and diverse sentences to describe concept relationships in everyday scenarios.

\section*{Limitations}

Our proposed DimonGen task involves generating several diverse sentences to describe the relationships between concepts. However, it does not take into account the number of relationships between different concept pairs. This can lead to problems when applying the model trained on the DimonGen dataset to other unseen concept pairs. For example, some concepts may have a small number of relationships, and asking the model to generate a greater number of diverse relationships may lead to \textit{hallucinations} which can be misleading when using the generative model for educational purposes. We leave this as a future work for the research community.

Additionally, the performance of the MoREE model is heavily dependent on the quality of the external corpora used in the retrieval stage. If the corpora do not contain any relevant information for the input concepts, the MoREE model will perform similarly to a vanilla MoE model. An alternative approach is to retrieve information from the Web~\citep{huang-etal-2022-understanding,lazaridou2022internet}.

Last, it should be noted that the base models used in this study were relatively small. Recent studies have demonstrated that large language models possess superior reasoning abilities compared to their smaller counterparts~\citep{wei2022chain,huang2022reasoning}. Future work on exploring the diversified generative commonsense reasoning ability of large language models is encouraged.

\section*{Acknowledgements}

This material is based upon work supported by the National Science Foundation IIS 16-19302 and IIS 16-33755, Zhejiang University ZJU Research 083650, IBM-Illinois Center for Cognitive Computing Systems Research (C3SR) and IBM-Illinois Discovery Accelerator Institute (IIDAI), gift grants from eBay and Microsoft Azure, UIUC OVCR CCIL Planning Grant 434S34, UIUC CSBS Small Grant 434C8U, and UIUC New Frontiers Initiative. Any opinions, findings, and conclusions or recommendations expressed in this publication are those of the author(s) and do not necessarily reflect the views of the funding agencies.

\bibliography{anthology,custom}
\bibliographystyle{acl_natbib}

\end{document}